\documentclass[letterpaper, 10 pt, conference]{ieeeconf}  

\IEEEoverridecommandlockouts                              

\usepackage{cite}
\usepackage{amsmath,amssymb,amsfonts}
\usepackage{algorithmic}
\usepackage{graphicx}
\usepackage{amsmath} 
\usepackage{textcomp}
\usepackage{xcolor}
\usepackage{booktabs}
\usepackage{siunitx}
\usepackage{tabularx}
\usepackage{multirow, makecell}
\usepackage{multicol}
\usepackage{arydshln}

\DeclareMathOperator*{\argmax}{argmax}

\title{\LARGE \bf
Human Scanpath Prediction in Target-Present Visual\\ Search with Semantic-Foveal Bayesian Attention
}

\author{João Luzio, Alexandre Bernardino, and Plinio Moreno
\thanks{João Luzio, Alexandre Bernardino, and Plinio Moreno are with the Institute for Systems and Robotics, Instituto Superior Técnico, University of Lisbon, Portugal. Email:   
{\tt\small joaoluzio14@tecnico.ulisboa.pt}}
\thanks{This work was supported by \textit{Fundação para a Ciência e Tecnologia}.}%
}

\begin{document}

\maketitle
\thispagestyle{empty}
\pagestyle{empty}

\begin{abstract}

In goal-directed visual tasks, human perception is guided by both top-down and bottom-up cues. At the same time, foveal vision plays a crucial role in directing attention efficiently. Modern research on bio-inspired computational attention models has taken advantage of advancements in deep learning by utilizing human scanpath data to achieve new state-of-the-art performance. In this work, we assess the performance of \textit{SemBA}-FAST, i.e. Semantic-based Bayesian Attention for Foveal Active visual Search Tasks, a top-down framework designed for predicting human visual attention in target-present visual search. \textit{SemBA}-FAST integrates deep object detection with a probabilistic semantic fusion mechanism to generate attention maps dynamically, leveraging pre-trained detectors and artificial foveation to update top-down knowledge and improve fixation prediction sequentially.  
We evaluate \textit{SemBA}-FAST on the COCO-Search18 benchmark dataset, comparing its performance against other scanpath prediction models. Our methodology achieves fixation sequences that closely match human ground-truth scanpaths. Notably, it surpasses baseline and other top-down approaches and competes, in some cases, with scanpath-informed models. These findings provide valuable insights into the capabilities of semantic-foveal probabilistic frameworks for human-like attention modelling, with implications for real-time cognitive computing and robotics. 
\end{abstract}

\section{Introduction}

Visual attention is influenced by both human ocular and cognitive systems. On the one hand, the human visual sensors (eyes) act as a hard-attention sensory mechanism \cite{activevision} that dynamically constrains the visibility of the content disposed across the visual field. On the other hand, the limitations imposed by the scarcity of brain resources, available for the visual-cognitive system, force the human active perception mechanism \cite{activeperception} to utilize the available information efficiently. 

The main anatomical constraint imposed by humans' visual sensing system, known as foveal vision \cite{fov}, effectively reduces the total amount of information to process during each gaze fixation. This is accomplished by overtly retaining a nitid central region of the field of view, known as the fovea, while progressively increasing the blur on its surrounding area, commonly referred to as the periphery. Simultaneously, a covert attention mechanism \cite{predvisualfix} processes the perceived information to determine the next gaze fixations, aiming for the most conspicuous regions in terms of task relevance.



Overall, human visual attention depends on two types of information signals: bottom-up and top-down. While bottom-up salient features can be directly extracted from visual stimuli alone \cite{predvisualfix}, top-down informed saliency is goal-directed, depending on the nature of the actual task at hand. Bottom-up cues \cite{trad_sal} are generally triggered by either color and brightness intensity contrasts or geometrical orientations of edges and patterns. Top-down guidance \cite{wolfe} derives from factors such as prior knowledge, task demands, expectations, and goals.

\begin{figure}
\includegraphics[width=\linewidth]{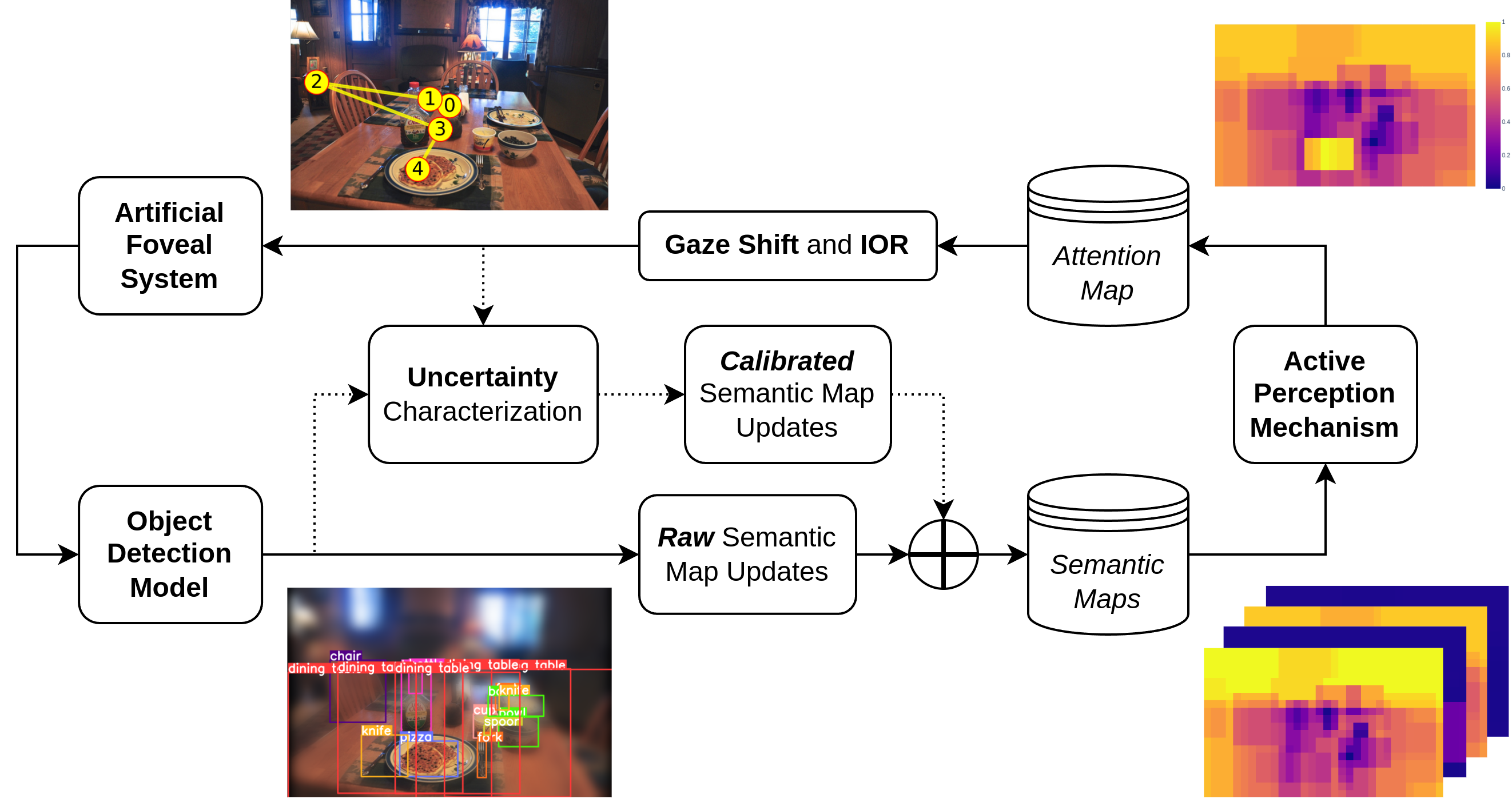}
\caption{Methodological pipeline of \textit{SemBA}-FAST \cite{ours}, inspired by \cite{main}. A foveal image is fed to a deep object detection model that generates multiple bounding-boxes and the respective categorical scores. This information is used to update world-fixed semantic maps. The updates can be performed with two approaches: using the raw scores of the object detector (\textbf{solid line}) or with scores calibrated according to the effects of the foveal image characteristics (\textbf{dotted line}), since increased blur in the periphery will increase uncertainty on the classification scores. Our active perception mechanism exploits the top-down information, gathered and stored in the semantic maps, to build a single attention map. We infer the next-best view from the generated attention map and apply an inhibition of return (IOR \cite{trad_sal,ivsn,irl}) mechanism to prevent revisiting previously searched locations. 
}
\label{fig:model}
\end{figure}

Among the many visual-cognitive tasks commonly performed by humans, two have been a source of particular interest \cite{sota} among the attention modelling research community: free-viewing and visual search. The former consists of freely exploring a scene \cite{main,kummerer} without any particular goal in mind. For this reason, free-viewing is mainly driven by bottom-up low-level features. The latter conforms to the process of searching for instances of a given target class in a visual field that may contain multiple distractors. Due to its goal-directed nature, visual search heavily depends on top-down features \cite{ivsn,transformers} such as scene context and semantics.

Visual search experiments are often conducted under two distinct setups \cite{cocosearch18}: target-present (TP), where the perceived visual stimulus contains at least one instance of the targeted class, and target-absent (TA), for the cases where no objects of the searched category are present across the field of view.

Understanding how humans search for targets by predicting their scanpaths (i.e. fixation sequences) has important implications for human-computer interaction and the development of systems that can anticipate user needs and intentions. Human gaze prediction can be of great value to a multitude of applications such as early diagnosis of cognitive impairments \cite{memory} and mental health conditions \cite{schizo}, as well as humanoid robotics \cite{activevision} and even foveated rendering \cite{fovrend}. 

Due to recent breakthroughs in deep learning models and the establishment of human fixation datasets and benchmarks \cite{cocosearch18,sota}, research on human attention prediction has moved from high-level feature extraction approaches (e.g. \cite{ivsn}) to models trained with actual human scanpath data (e.g. \cite{irl,ffm,kummerer}). Although models like Gazeformer \cite{gazeformer}, HAT \cite{transformers}, and CLIPgaze \cite{clipgaze} considerably improved the state-of-the-art performance, they end up lacking the biological plausibility and explainability of methods (e.g \cite{main}) that try to mimic the visual-cognitive system's behavior. For instance, human beings do not learn how to explore their surrounding environment or complete complex visual tasks from the experiences and data of fellow humans. Indeed, despite the need for top-down knowledge such as semantics and context \cite{predvisualfix}, human infants are not informed with any type of scanpath data when learning how to explore complex crowded scenes to swiftly find target objects. In fact, humans intuitively learn such skills \cite{wolfe}, as our built-in visual-cognitive system effectively guides our attention and allows us to better perceive and interact with the surrounding environment. 

After IVSN \cite{ivsn}, a zero-shot model that builds attention maps mainly from extracted top-down features, not many pure stimulus-driven models have been published. Inspired by the anatomical properties of the human eye and the brain's attention map assembling process \cite{trad_sal} we proposed a semantic-based Bayesian attention methodology \cite{ours} which we appropriately named \textit{SemBA}. We followed the common trend in neurology literature \cite{wolfe,foundations} by combining foveal vision and semantic data fusion in a single architecture.

In this work, our focus is on TP visual search. We aim to assess whether our proposed top-down probabilistic attention framework for foveal active search tasks, i.e. \textit{SemBA}-FAST \cite{ours}, can accurately predict human-generated scanpaths. Both our model and methodology, highlighted in Fig. \ref{fig:model}, are inspired by the work of Dias et al. \cite{main} on semantic-foveal active perception for scene exploration (i.e. free-viewing). The main contributions of this specific work are the following:

\begin{itemize}

    \item We establish a comparison between \textit{SemBA}-FAST's \cite{ours} experimental results and the results attained by other state-of-the-art models \cite{ivsn,irl,vqa,gazeformer,ffm,transformers,clipgaze} on a well-known visual search benchmark, namely the COCO-Search18 \cite{cocosearch18} dataset, using off-the-shelf evaluation metrics \cite{sota}.

    \item As a pure semantic-based model, \textit{SemBA}-FAST has been shown to attain a cumulative performance in TP visual search tasks comparable to human results \cite{cocosearch18}.

    \item We experimentally demonstrate that \textit{SemBA}-FAST can confidently outperform baseline models and other top-down approaches \cite{ivsn}. Moreover, the studied model can also surpass human fixation data-informed models \cite{irl,vqa,gazeformer,ffm,transformers,clipgaze} with regard to some fixation prediction metrics.
    
\end{itemize}

\section{Background \& Related Work}

\subsection{Deep Object Detection}

Object recognition is a well-known computer vision problem that consists of localizing (determining the positions through bounding-boxes) and categorizing (determining the corresponding class) visible objects in images and video frames. Initially dominated by traditional algorithms with several limitations \cite{odreview}, the field has undergone a significant transformation with the evolution of deep learning models. 

In the object detection research field \cite{odreview}, models are divided into two-stage detectors and single-stage detectors. Two-stage detectors (e.g. R-CNN \cite{rcnn}) first propose regions of interest and only then proceed to classify these areas. In contrast, single-stage detectors (e.g. YOLO \cite{yolo_og} and its subsequent versions, i.e. YOLOv2 to YOLOv11 \cite{yolo_rev2}) omit the search phase for regions of interest and proceed directly to predicting the location and category of objects, which tends to increase both train and inference speeds. More recently, transformer-based architectures (e.g. DETR \cite{detr}) have disrupted the state-of-the-art on object recognition, treating the task as a direct ensemble prediction problem.

In general, single-stage detectors have been shown to outperform two-stage detectors in terms of speed and, increasingly, accuracy. The YOLO family \cite{yolo_rev1} has evolved continuously, becoming one of the most popular and effective classes of detectors for real-time object recognition. Despite all the developments \cite{odreview}, challenges such as detection in variable lighting conditions and object occlusion still persist.

\subsection{Attention Modelling}

Comprehending how humans control and direct their gaze movements \cite{predvisualfix} is fundamental for understanding our attention-guided visual behavior patterns. While visual-attention prediction has been a source of curiosity for neurologists and psychologists alike for many decades \cite{wolfe,foundations}, this topic has only recently awakened the interest of both computer vision and deep learning research communities. Foundational work \cite{trad_sal} on low-level feature extraction for visual saliency mapping, carried out by L. Itti and C. Koch, fostered many posterior developments \cite{benchmark} in bottom-up saliency analysis for human attention modelling.

Besides overlooking top-down task-specific features \cite{transformers}, pure bottom-up saliency-based attention research has solely focused on distinguishing between the conspicuity levels of different regions across the entire field of view. While conspicuity maps may effectively model the spatial distribution of gaze fixations, they do provide little to no insights into the sequential order of predicted focal points. For these reasons, human scanpath prediction poses a real challenge to traditional attention modelling, as we are interested in predicting not only fixation locations but also fixation temporal order. 

\subsection{Human Scanpath Prediction}

We already asserted that goal-directed gaze prediction relies on the ability to extract and process top-down features. Zhang et al. \cite{ivsn} heavily enriched biologically inspired scanpath prediction research with the proposal of their top-down zero-shot model, referred to as the Invariant Visual Search Network (IVSN). Zero-shot search consists of finding instances of classes that did not appear in the training dataset.

After IVSN, a multitude of deep-learning-based human scanpath prediction models \cite{sota} was propelled with the establishment of the first large-scale visual search benchmark dataset, known as COCO-Search18 \cite{cocosearch18}. Taking advantage of eye-tracking technology, COCO-Search18 provides many fixation sequences, generated by 10 human subjects, under both target-present and target-absent settings. By incorporating scanpaths in the training process, models such as IRL \cite{irl} (inverse reinforcement learning) and FFMs \cite{ffm} (foveated feature maps) have been able to surpass the performances of IVSN \cite{ivsn} and other state-of-the-art baseline models \cite{sota}. Chen et al. \cite{vqa} has also been successful in expanding inverse reinforcement learning-based scanpath prediction to another demanding task, namely visual question answering (VQA). More recently, Gazeformer \cite{gazeformer} and HAT \cite{transformers} have been able to leverage soft-attention mechanisms, characteristic of transformer-based architectures (ViT), to outperform all their competitors, both in TP and TA visual search tasks. Both models draw inspiration from the human visual system, implementing a dynamic visual memory through the simulation of a simplified foveated retina. Finally, CLIPgaze \cite{clipgaze} utilizes a large vision language model (VLM), namely CLIP, to extract and provide pre-matched features for images and their respective target prompts, to search for objects under TP, TA, and also zero-shot settings. 

\section{Methodology} \label{methodology}

In this section, we describe the full pipeline for our human attention prediction methodology: \textit{SemBA}, i.e. Semantic-based Bayesian Attention. We consider a mechanism \cite{ours} for top-down semantic information fusion, within a probabilistic framework, which leverages pre-trained object detectors. We exhibit a schematic illustration of our approach in Fig. \ref{fig:model}. 

\subsection{Spatial Constraints}

Similar to other state-of-the-art human scanpath prediction methodologies \cite{irl,gazeformer,transformers}, we consider two-dimensional images as fixed fields of view, where the spatial configuration and boundaries of the scene cannot be dynamically changed. Moreover, we assume that the visual field has a static configuration \cite{ours}, where the disposition of the objects contained in it does not vary across time. Taking this assumption into consideration, we define each fixation's coordinates at the pixel level, to mitigate precision loss associated with grid discretization \cite{transformers}. Consider an image of dimensions $\textit{height} \times \textit{width}$ and a given initial fixation $f_0$, commonly set in the center of the visual field. Our model generates a sequence of human-like fixation points $f_1, f_2, \dots, f_n$, where each fixation $f_t, \forall t \in \{ 1, \dots, n\}$ corresponds to a specific pixel location within the image.  Each fixation sequence length $n$ varies from scene to scene since it depends on the established termination criteria \cite{ivsn}. For \textit{SemBA}'s output semantic and attention maps (see Fig. \ref{fig:model}), we select a typical two-dimensional Cartesian representation that spatially encodes the surrounding environment, in the form of a $Y \times X$ context grid. Through action space discretization, we aim at reducing computational costs \cite{cocosearch18} and approximate the human cognitive system's sensitivity levels \cite{foundations} by processing broader regions of interest instead of singular pixels.

\subsection{Probabilistic Framework}

As depicted in Fig. \ref{fig:model}, each detection consists of a bounding-box and a score vector $S$ of dimension $K$, which is typically normalized. A vector $S$ contains a set of confidences, each associated with the presence of an instance of a particular known class inside the limits of its bounding-box:
\begin{equation}
    S = \left( s_1, s_2, \thinspace \dots \thinspace,  s_K \right), \thinspace 0\leq s_{k} \leq 1, \forall k \in \mathcal{C}
    \label{eq:scores}
\end{equation}
\noindent where $\sum_{i=1}^K s_{i} \!=\! 1$ and $\mathcal{C} \!\subseteq\! \left\{1,\dots,K \right\}$ represents the set of classes known by the detector.
The category of an object located in $\mathbf{x} = (x,y)$ is modeled as a Dirichlet-compound multinomial distribution (belief) with parameters $\boldsymbol{\beta}^\mathbf{x} \in \mathbb{R}^K$

\begin{equation}
    P \left( C^{\mathbf{x}} = k \mid \boldsymbol{\beta}^{\mathbf{x}} \right) = \dfrac{n!}{\left(\sum_{i=1}^{K} \beta_{i}^{\mathbf{x}}\right)^n} \prod\limits^K_{i=1} \dfrac{\left(\beta_{i}^{\mathbf{x}}\right)^{n_i}}{n_i !}
    \label{eq:semanticmap}
\end{equation} where $n\!=\!\sum_{i=1}^{K} n_i, n_k \!=\! 1$ and $n_i \!=\! 0, \forall i \! \in\! \mathcal{C} \setminus k$ \cite{ours}. We build semantic maps for each class $k \in \mathcal{C}$, as depicted in Fig. \ref{fig:model}, through an extension of \eqref{eq:semanticmap} for all grid locations $\mathbf{x}$. 

The parameters of the Dirichlet belief's set are initialized as $\beta^\mathbf{x}_k = 1, \forall k \in \left\{ 1, \dots, K \right\}$, to define a flat Dirichlet distribution that corresponds to a non-informative prior. 

Essentially, the goal of our methodology is to update the Dirichlet-compound multinomial distribution's $\boldsymbol{\beta}^\mathbf{x}$ parameters using the scores \eqref{eq:scores} from new observations, which are sequentially extracted from every fixation along the scanpath. 

\subsection{Semantic Data Fusion}

The most straightforward way to update the semantic maps is to use the classifier output score vectors, as conveyed by the solid line path in Fig. \ref{fig:model}. To encompass distinct scores obtained across multiple fixations, we apply a classifier fusion rule \cite{kaplan}, developed from a subjective logic perspective
\begin{equation}
    \beta_{k}^{\mathbf{x}} \longleftarrow \dfrac{\beta_{k}^{\mathbf{x}} \left( 1 + \dfrac{\lambda_{k}}{\sum_{j=1}^{K} \beta_{j}^{\mathbf{x}} \lambda_{j}} \right)}{1 + \dfrac{\min_i \lambda_{i}}{\sum_{j=1}^{K} \beta_{j}^{\mathbf{x}} \lambda_{j}}} , \forall k \in \mathcal{C}
    \label{eq:kaplanrule}
\end{equation}
where $\lambda_k$ represents the observation's categorical likelihood, i.e. $\lambda_k = P(S \mid C = k)$, for each class $k \in \mathcal{C}$. This update rule is borrowed from Kaplan's work \cite{kaplan} on classifier fusion.
We use a score vector likelihood $P (S \mid C = k )$ to update the Dirichlet priors $\boldsymbol{\beta}^\mathbf{x}$ if its bounding-box overlaps the region that corresponds to the grid's $\mathbf{x}$ spatial coordinates.

\subsection{Foveal Calibration} \label{fovcalib}

If we assume that a detector is indeed well-calibrated \cite{uncertainty} then $s_k = P(C = k \mid S) \propto P(S \mid C = k) = \lambda_k$, according to Bayes' rule, and therefore the actual raw scores \eqref{eq:scores} can be directly applied to Kaplan's rule \eqref{eq:kaplanrule} to perform new updates.

However, classifiers are often not properly calibrated to reflect the posterior probabilities \eqref{eq:semanticmap} of the actual class, contrary to what is frequently assumed in many classification problems \cite{uncertainty}. To attain the likelihoods $\lambda_k$ we need to learn a proper sensor model $P(S \mid C = k)$ from a large dataset. 

Moreover, we have to account for the fact that the detector has been pre-trained with a regular image dataset and therefore is susceptible to the effects of peripheral distortion introduced by the incorporation of the artificial foveal system \cite{fovsys}. According to the anatomical properties of the human eye \cite{fov}, the amount of blur in a given region of the field of view is directly related to its distance to the center of the fovea. Hence, we adapt the sensor model to accommodate the characteristics of the fovea, by dividing the field into $D$ discrete distance levels, radiating from the center of the fovea (i.e. focal point). Categorical likelihoods can henceforth be posed as $P(S \mid C = k, d)$, according to the distance level $d \in \{ 1, \dots, D\}$ in which the respective bounding-box falls.          

Assuming that each $P(S \mid C = k, d)$ can be modeled as a Dirichlet, we need to learn $K \times D$ Dirichlet distributions, one for each class and distance level, with the corresponding Dirichlet parameters $\boldsymbol{\alpha}_{k,d} \in \mathbb{R}^K$. The Dirichlet likelihoods \cite{bishop} for each class $k$ and distance level $d$ are estimated as
\begin{equation}
    \mathrm{Dir} \! \left( S \mid \boldsymbol{\alpha}_{k,d} \right) =  \dfrac{\Gamma \! \left( \sum^K_{i=1} \alpha_{k,d,i} \right)}{\prod^K_{i=1} \Gamma\!\left(\alpha_{k,d,i}\right)} \prod\limits^K_{i=1} s_{i}^{\alpha_{k,d,i} - 1}
    \label{eq:dirichlet}
\end{equation}
\noindent involving Euler's Gamma function: $\Gamma(z) =  \int_0^\infty t^{z-1} e^{-t} dt$.

\begin{figure}
\centering
\includegraphics[width=0.92\linewidth]{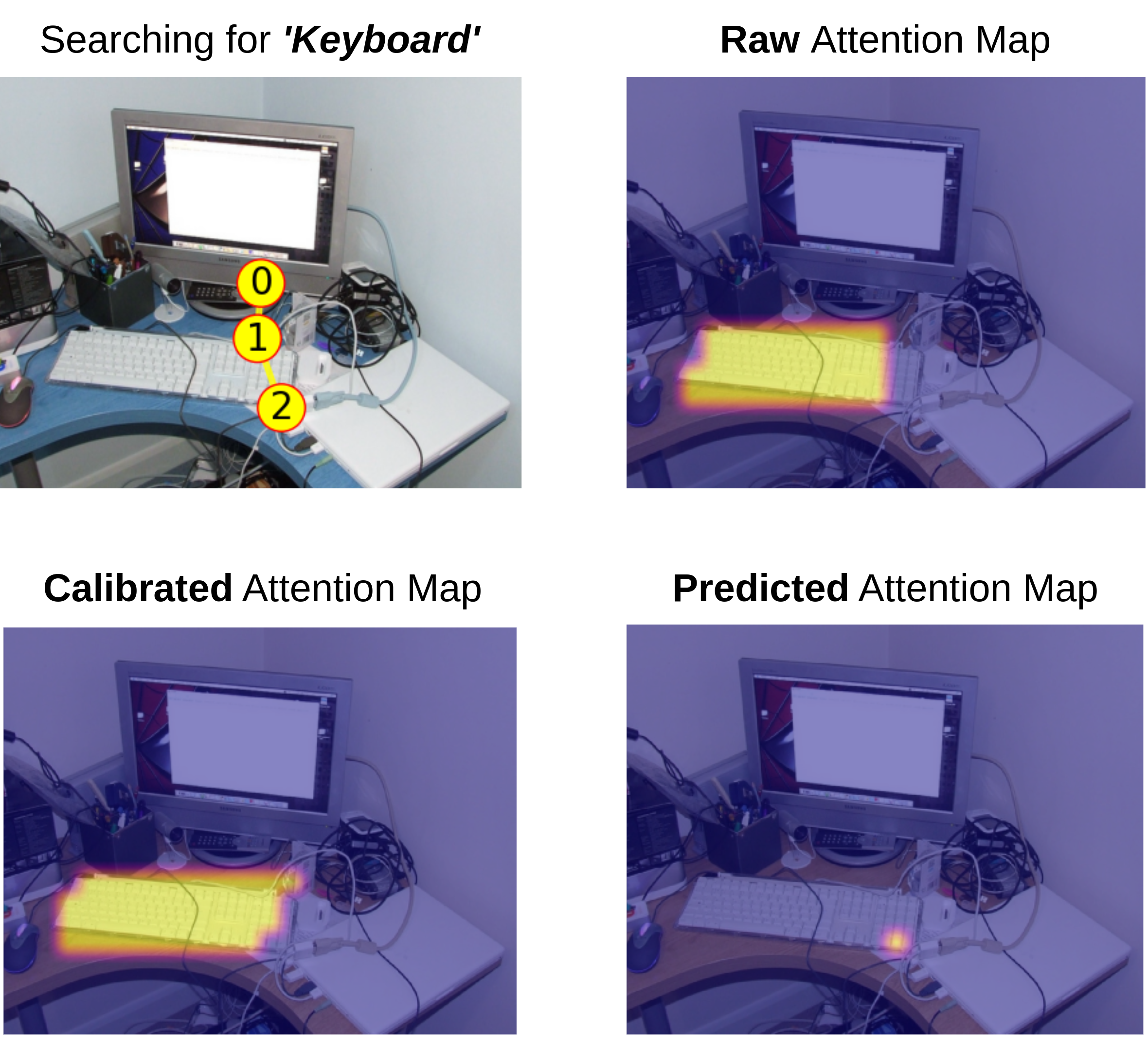}
\caption{Illustrative sample case containing attention (heat) maps generated by distinct approaches in a target-present visual search setup. In this particular example, the targeted class is "\textit{keyboard}" and we present the map states after 2 iterations. The raw, calibrated and predicted attention maps are generated and used by \textit{SemBA}-FAST \textit{Base}, \textit{Calib}, and \textit{Pred}, respectively.
}
\label{fig:compmaps}
\end{figure}

Objectively, each set of Dirichlet parameters is estimated by fitting multiple detection data, consisting of normalized score vectors \eqref{eq:scores}, according to the respective ground-truth class $k \in \mathcal{C}$ and focal distance level $d \in \{ 1, \dots, D \}$. Given the fact that there is no known closed-form maximum-likelihood estimation for Dirichlet distributions, we consider a simple iterative method, proposed by T. Minka \cite{dirichlet}, to fit semantic data, extracted from a large dataset, to a Dirichlet likelihood \eqref{eq:dirichlet}. With this technique, we generate the $\boldsymbol{\alpha}_{k,d}$ sets with which we capture the aleatoric uncertainty, derived from data-related factors (e.g. illumination and occlusion), and the additional uncertainty that is imposed by the foveal sensor.

Following the dotted path in Fig. \ref{fig:model}, we can utilize the new calibrated likelihoods $\mathrm{Dir} \! \left( S \mid \boldsymbol{\alpha}_{k,d} \right) \propto P(S \mid C = k) = \lambda_k$ to update the beliefs $\boldsymbol{\beta}^\mathbf{x}$, while keeping the fusion rule \eqref{eq:kaplanrule}.

\subsection{Active Perception}

To yield the predicted attention map, for each fixation, we finally invoke our active perception mechanism. Essentially, to build these saliency-like maps, we exploit the information cumulatively gathered in the semantic maps. Assuming that, for a given visual search task, the goal is to find an instance of a target class $k$, we assemble a probabilistic map according to the posterior probabilities \eqref{eq:semanticmap}, extracted from the $k$-th semantic map. Then, we greedily determine $\mathbf{x}^\ast$, the region that exhibits the highest posterior, as the next-best fixation:
\begin{equation}
    \mathbf{x}^\ast = \argmax\limits_{\mathbf{x}} \thinspace P \left( C^{\mathbf{x}} = k \mid \boldsymbol{\beta}^{\mathbf{x}} \right)
    \label{eq:gazeselect}
\end{equation}

We iteratively apply this decision mechanism to select the next fixation $f_{t+1} \equiv \mathbf{x}^\ast$ (in pixels), according to the information collected across all previous fixation points $f_{0:t}$, until the target object is eventually found. On top of this, we also apply inhibition of return (IOR \cite{ivsn,irl}) to prevent redirecting the gaze toward previously visited locations.

\begin{figure}
\centering
\includegraphics[width=0.95\linewidth]{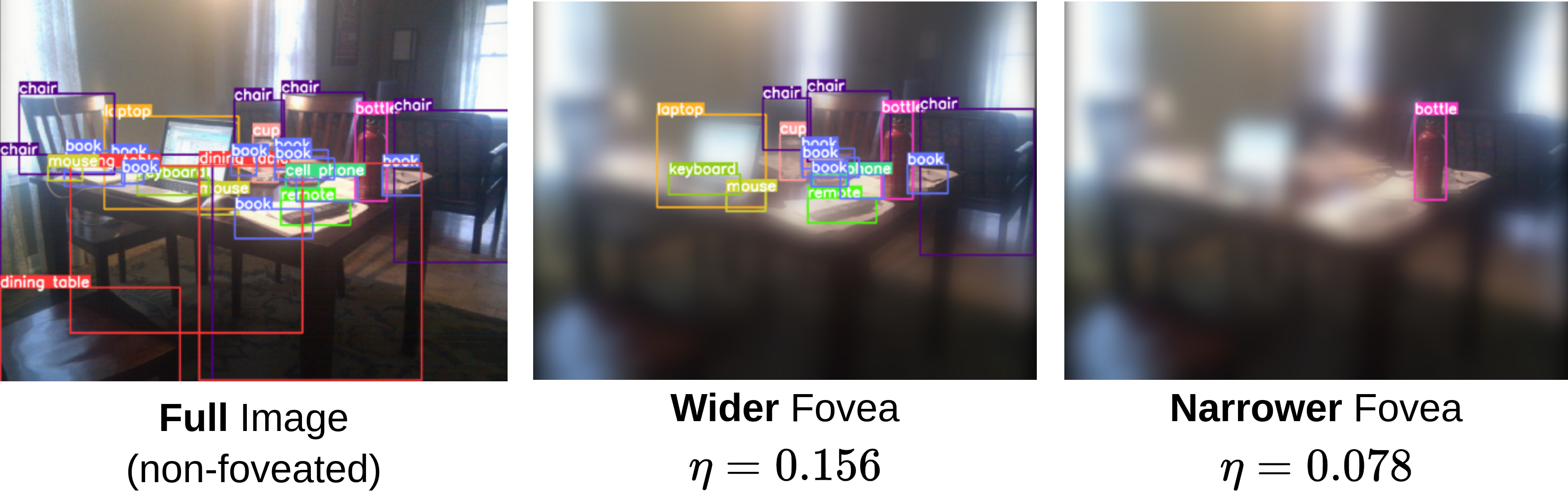}
\caption{Peripheral distortion effects on object detection (with YOLOv5 \cite{yolo_rev2}). Narrower fovea dimensions hinder the detector's capabilities in terms of localizing and classifying objects in broader regions around focal points.}
\label{fig:fov_shapes}
\end{figure}

Finally, we define three different approaches that derive from our methodological pipeline. The sole distinction between each approach lies in the information and technique used to update the beliefs $\boldsymbol{\beta}^\mathbf{x}$, which ground the posterior \eqref{eq:semanticmap}. Fig. \ref{fig:compmaps} illustrates the differences between generated maps.

\begin{figure}
\includegraphics[width=\linewidth]{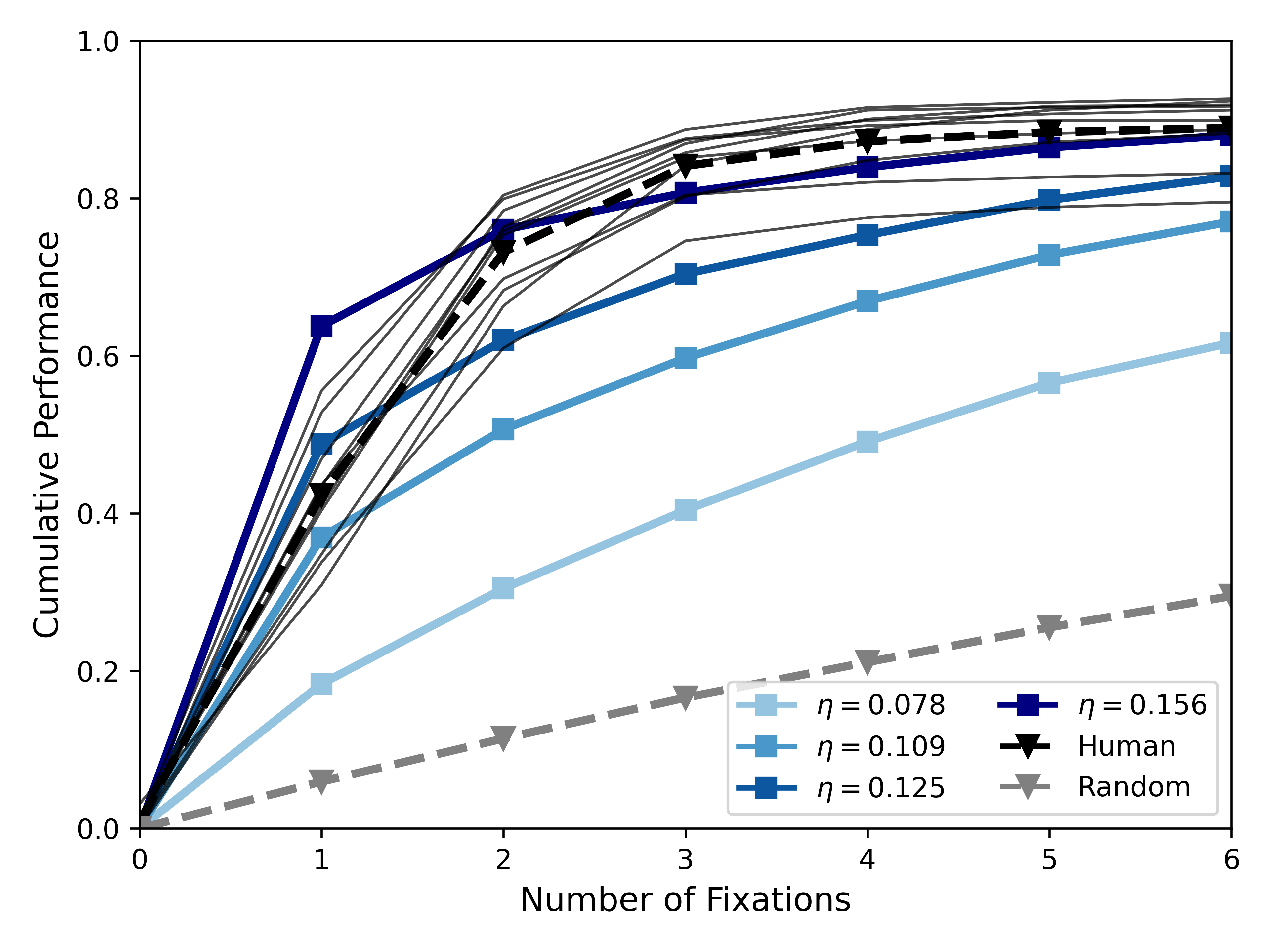}
\caption{Comparison of the cumulative performances of \textit{SemBA}-FAST \textit{Pred} on the test partition of COCO-Search18 \cite{cocosearch18} under different foveal dimensions. Augmenting the value of $\eta$ implies widening the fovea. We also present the average performances of humans and random selection. The thin lines correspond to the individual performances of human subjects.}
\label{fig:fov}
\end{figure}

\subsubsection{\textbf{\textit{SemBA}-FAST's Baseline Approach}}
For our first and simplest approach, we na\"ively assume that the sensor's output scores of each detection are well enough calibrated \cite{uncertainty} to reflect the actual categorical likelihoods, i.e. $\lambda_k = s_k, \forall k \in \mathcal{C}$. This procedure is represented by the solid path in Fig. \ref{fig:model}, where scores \eqref{eq:scores} are directly fused \eqref{eq:kaplanrule} into semantic maps. We refer to this approach as \textit{SemBA}-FAST \textit{Base}.

\subsubsection{\textbf{\textit{SemBA}-FAST's Calibrated Approach}}

Following the sensor calibration procedure described in Section \ref{fovcalib}, we update \eqref{eq:kaplanrule} the semantic maps with the Dirichlet likelihoods \eqref{eq:dirichlet} of the output scores, i.e. $\lambda_k = \mathrm{Dir} \! \left( S \mid \boldsymbol{\alpha}_{k,d} \right), \forall k \in \mathcal{C}$, instead of the original raw sensor scores \eqref{eq:scores}. We set the peripheral distortion level $d$ according to each detection's distance to the respective fixation point (dotted path drawn in Fig. \ref{fig:model}). We refer to this approach as \textit{SemBA}-FAST \textit{Calib}.

\subsubsection{\textbf{\textit{SemBA}-FAST's Predictive Approach}}

We kick-start our predictive approach with a simple question: What would be the expected value for the detector's output if the agent were to next place its gaze on top of each single location $\mathbf{x}$?

Following \cite{ours}, we exploit the properties of the expectation operator to determine the expected value of the sensor output if centered ($d=1$) upon the object (possibly) located in $\mathbf{x}$:
\begin{equation}
    \Bar{S}^{\mathbf{x}} = \mathbb{E} \left[ S^{\mathbf{x}} \mid \boldsymbol{\beta}^{\mathbf{x}}, d \!=\! 1\right]  = \sum\limits^K_{k=1} \dfrac{\boldsymbol{\alpha}_{k,1}}{\sum^K_{j=1} \alpha_{k,1,j}} \dfrac{\beta^{\mathbf{x}}_{k}}{\sum^K_{j=1} \beta^{\mathbf{x}}_{j}}
    \label{eq:expect}
\end{equation}

We build our attention map after simulating multiple semantic map updates \eqref{eq:kaplanrule}, more specifically one for each map location $\mathbf{x}$, fusing the expected sensor output scores \eqref{eq:expect} with the current beliefs $\boldsymbol{\beta}^\mathbf{x}$. Similarly to \textit{SemBA}-FAST \textit{Base}, we update the base beliefs $\boldsymbol{\beta}^\mathbf{x}$ directly with the raw scores \eqref{eq:scores}. It is important to note that we only keep track of the base beliefs, updated with actual detections, while the expected maps are discarded after the fixation $f_{t+1}$ is predicted. We henceforth refer to this approach as \textit{SemBA}-FAST \textit{Pred}.

\section{Experiments}

\begin{table*}
\centering
\caption{Comparison of Human Scanpath Prediction Models using Fixation Sequence and Next Fixation Evaluation Metrics}
\begin{tabular}{cccccccccccc} \toprule
    \multirow{2}{*}{\hfil \parbox{2.1cm}{\textbf{Human Scanpath Prediction Model}}} & \multicolumn{5}{S}{\textbf{Scanpath (Fixation Sequence)}} & & \multicolumn{3}{S}{\textbf{Next Fixation Prediction}} & & \multirow{2}{*}{\textbf{Time Cost}$^{\mathrm{1}}$ $\downarrow$}\\ \cmidrule{2-6} \cmidrule{8-10}
    & {\textbf{SS} $\uparrow$} & {\textbf{FED} $\downarrow$} & {\textbf{SemSS} $\uparrow$} & {\textbf{SemFED} $\downarrow$} & {\textbf{SR} $\uparrow$} & & {\textbf{cNSS} $\uparrow$} & {\textbf{cIG} $\uparrow$} & {\textbf{cAUC} $\uparrow$} & & \\ \midrule
    {Human Consistency}  & 0.463 & 2.353 & 0.470 & 2.144 & 0.786 & & - & - & - & & - \\
    {Random Gaze Selection}  & 0.278 & 4.813 & 0.260 & 4.823 & 0.196 & & - & - & - & & - \\
    \midrule
    {IVSN \cite{ivsn} Random Target} & 0.297  & 4.009 & 0.330 & 3.711 & 0.617 & & 0.774 & -0.241 & 0.713 & & 0.154\\
    {IVSN \cite{ivsn} Picked Target} & 0.346 & 3.360 & 0.380 & 3.034 & 0.673 & & 0.805 & -0.227 & 0.743 & & 0.187 \\
    \midrule
    {IRL \cite{irl}} & 0.416  & 2.757 & 0.471 & 2.321 & 0.730 & & 1.977  & -9.709 & 0.913 & & 0.177 \\
    {Chen \textit{et al.} \cite{vqa}} & 0.451  & 2.187 & 0.470 & 1.898 & - & & 2.606  & -1.273 & 0.956 & & 0.803 \\
    {FFMs \cite{ffm}} & 0.392 & 2.693 & 0.407 & 2.425 & - & & 2.376 & 1.548 & 0.932 & & 0.277 \\
    {Gazeformer \cite{gazeformer}} & \textbf{0.504} & 2.072 & 0.490 & 1.928 & - & & - & - & - & & \textbf{0.141} \\
    {HAT \cite{transformers}} & 0.468  & 2.063 & 0.540 & 1.522 & - & & \textbf{5.086}  & \textbf{2.399} & \textbf{0.977} & & - \\ 
    {CLIPgaze \cite{clipgaze}} & 0.476  & \textbf{2.014} & \textbf{0.545} & \textbf{1.489} & - & & -  & - & - & & - \\ \midrule
    {\textit{SemBA}-FAST \cite{ours} \textit{Base} (ours)} & 0.409 & 2.639 & 0.429 & 2.285 & 0.740 & & 2.375 & 0.320 & 0.914 & & 1.000 \\
    {\textit{SemBA}-FAST \cite{ours} \textit{Calib} (ours)} & 0.406 & 2.732 & 0.422 & 2.408 & 0.700 & & 2.252 & 0.322 & 0.910 & & 1.054\\
    {\textit{SemBA}-FAST \cite{ours} \textit{Pred} (ours)} & 0.413 & 2.616 & 0.431 & 2.272 & \textbf{0.742} & & 2.375  & 0.326 & 0.914 & & 1.694\\ \bottomrule
    
    \multicolumn{12}{l}{
    $^{\mathrm{1}}$Per iteration, relative to \textit{SemBA}-FAST \textit{Base} fed with $320 \times 512$ pixel scale input images. 
    }
    \label{tab:main}
\end{tabular}
\end{table*}

As asserted in Section \ref{methodology}, our methodology relies heavily on accurate 2D object localization and classification. Hence, we incorporate YOLO \cite{yolo_og}, a state-of-the-art deep detection algorithm, in our methodology to generate object predictions. 

Despite the increased performance achieved by more recent and sophisticated YOLO versions \cite{odreview}, we have elected YOLOv5 as our go-to model, primarily due to its remarkable efficiency and reliability. Overall, YOLOv5 is able to yield object predictions that, to a degree, are not much less accurate than the ones yielded by its successors \cite{yolo_rev1}. YOLOv5 is also quite fast and has been extensively reviewed \cite{yolo_rev1,yolo_rev2}.

\subsection{Datasets}

We conducted our evaluation in the test partition of the benchmark dataset COCO-Search18 \cite{cocosearch18}. COCO-Search18 is a visual search dataset that includes human scanpaths from 10 human subjects, performing search tasks that involve 18 distinct classes of target objects. The dataset is split into two search tasks: target-present and target-absent, each comprising 3101 images. COCO-Search18 is a subset of the large-scale image dataset COCO 2017 \cite{coco}. For this reason, we trained YOLOv5 with all the COCO 2017 images (for 80 classes), excluding the 6202 images from COCO-Search18. For the evaluation phase, we fed YOLOv5 with instances of COCO-Search18 on their maximum available scale, i.e. 1050x1680, contrasting with Gazeformer \cite{gazeformer} and HAT \cite{transformers}, which rescale inputs to 640x1024 and 320x512, respectively.

\subsection{Experimental Setup}

To simulate a foveal sensor, we apply Almeida's artificial foveal system \cite{fovsys}, centering the fovea around each selected focal point. Instead of a fixed circular foveal shape \cite{fov}, we define the foveal dimensions to be aligned with each image axis length, such that $\sigma_0^x = \eta \times \textit{width}$ and $\sigma_0^y = \eta \times \textit{height}$, therefore allowing for elliptic shapes, as those exhibited in Fig. \ref{fig:fov_shapes}. By taking this approach, we define an adaptive fovea that can be applied to any image, regardless of its $\textit{height} \times \textit{width}$ scale, and with which we try to maintain the proportions of observed objects between scenes. Nevertheless, this property is still unwarranted, as it hinges on multiple factors that are intrinsic to the scene's apparatus itself. We define $\eta = 0.156$, as it led \textit{SemBA}-FAST's \textit{Pred} variant (see Fig. \ref{fig:fov}) to achieve cumulative performances similar to those attained by humans in COCO-Search18 \cite{cocosearch18}.

We use the same set of images considered for YOLOv5's training procedure (each foveated around a random focal point) to train our calibration mechanism, fitting the parameters of multiple (i.e. $K \times D$, where $K=80$ classes) Dirichlet distributions to detections generated by YOLOv5. Following the description in Section \ref{fovcalib}, we spatially divide the scene into $D = 7$ uniform peripheral distortion levels, according to the distance to the center of the fovea. To find the level to which a given detection should be associated, we compute the Mahalanobis distance \cite{mahalanobis} between the center of its bounding-box and the focal point. By selecting the Mahalanobis distance as our peripheral distortion level decision metric, we can easily adapt to possible variations in the foveal shape, as long as its covariance matrix is built according to $\sigma_0^x$ and $\sigma_0^y$, based on input's \textit{width} and \textit{height}.


To promote a proper and fair comparison, we follow \cite{irl,transformers} and set the maximum length of each predicted scanpath (i.e. truncate) to 7 fixations \cite{cocosearch18}, including $f_0$ which is located in each image's center. 
Moreover, as proposed in \cite{transformers}, we set \textit{SemBA}-FAST's attention map resolution to $20 \times 32$, similarly to Chen \textit{et al.} \cite{vqa}, IRL \cite{irl}, FFMs \cite{ffm}, and Gazeformer \cite{gazeformer}. Following IRL \cite{irl}, we also apply an inhibition of return (IOR, also referred to as inhibitory spatial tagging) mechanism in a $3 \times 3$ grid around each focal point. 
Finally, and similarly to the choice of IVSN's proponents \cite{ivsn}, we also utilize an \textit{Oracle} that terminates the search once the gaze is fixed upon the targeted object. By incorporating such an \textit{Oracle}-based strategy, we circumvent the thought challenge of defining a success (i.e. termination) criterion.

\subsection{Evaluation Metrics} \label{sec:metrics}

To assess the performance of human scan path prediction models, it is important to distinguish between two types of evaluation metrics \cite{transformers}: 1) scanpath metrics, which assess the similarity between human and generated fixation sequences, and 2) next fixation metrics, based on the level of conspicuity that a generated attention map attributes to the next ground-truth (human) fixation, given all the previous fixations. We also report the average computational time costs per iteration. 

\subsubsection{Scanpath Metrics}

Sequence Score (\textbf{SS}) converts scanpaths into sequences of fixation cluster IDs and compares them using the Needleman-Wunsch string matching algorithm \cite{needlewunsch}, originally developed to evaluate the similarity of amino acid sequences between proteins. Fixation Edit Distance (\textbf{FED}) also transforms scanpaths into strings of fixation cluster IDs, but it uses Levenshtein's algorithm \cite{levenshtein} to measure the dissimilarity between them. Semantic Sequence Score (\textbf{SemSS}) differs from SS \cite{ffm} by comparing strings of categorical (semantic) labels, corresponding to the sequence of fixated objects, rather than cluster IDs, while still utilizing the Needleman-Wunsch algorithm. Similarly, Semantic Fixation Edit Distance (\textbf{SemFED}) \cite{gazeformer} compares semantic strings but employs Levenshtein's algorithm, akin to FED.  We utilize the listed metrics to compare each model's generated fixation sequence with the human ground-truth scanpaths. Finally, Scanpath Ratio (\textbf{SR}) \cite{irl} is computed as the Euclidean distance between the initial fixation location (i.e. image's center) and the location of the target (i.e. bounding-box's center) divided by the summed Euclidean distances between all scanpath's focal points (i.e. fixations).


\subsubsection{Next Fixation Metrics}

Most scanpath prediction models only generate a single attention map for a given scene (e.g \cite{irl,gazeformer}). This attention map (which can also be interpreted as a saliency map) is then typically iteratively explored, via a sequential selection of the most conspicuous regions (as focal points) with an IOR mechanism \cite{ivsn}. To assess the quality of generated attention (i.e. saliency) maps \cite{benchmark}, intensities of ground-truth fixated locations (from human scanpaths) can be extracted from the maps to compute performance metrics.

Normalized Scanpath Saliency (\textbf{NSS}) measures the average intensity of fixated cells (from a given scanpath) in a saliency map that has been normalized to have zero mean and unit variance.
Information Gain (\textbf{IG}) compares the saliency map, converted into a probability distribution, to a baseline model by evaluating the average log-probability of fixated cells. Following \cite{transformers}, we consider a baseline fixation density map, created by averaging the smoothed density maps (with Gaussian kernels of one degree of visual angle) of all fixations from the COCO-Search18 training set \cite{cocosearch18}. Area Under the Curve (\textbf{AUC}), more specifically AUC-Judd \cite{auc}, measures how well a saliency map predicts human fixations by calculating the area under the well-known ROC curve. Essentially, this metric compares the intensities at ground-truth fixated regions to ground-truth non-fixated regions.

Recent works \cite{ffm,transformers} introduced the concept of conditional saliency metrics (e.g. \textbf{cNSS}, \textbf{cIG}, \textbf{cAUC}). Conditional metrics assess how accurately a model-generated saliency map can predict the next ground-truth fixation, granting the model access to the fixation history of the scanpath under evaluation. This is extremely relevant for models such as FFMs \cite{ffm}, HAT \cite{transformers}, and \textit{SemBA}-FAST (ours) which, similarly to the visual-cognitive system \cite{wolfe}, iteratively update the attention map. Following our methodological pipeline, newly acquired information is fused with cumulatively gathered knowledge \cite{activeperception}, obtained from previous fixations.

\subsection{Other Models}

\begin{figure}
\includegraphics[width=\linewidth]{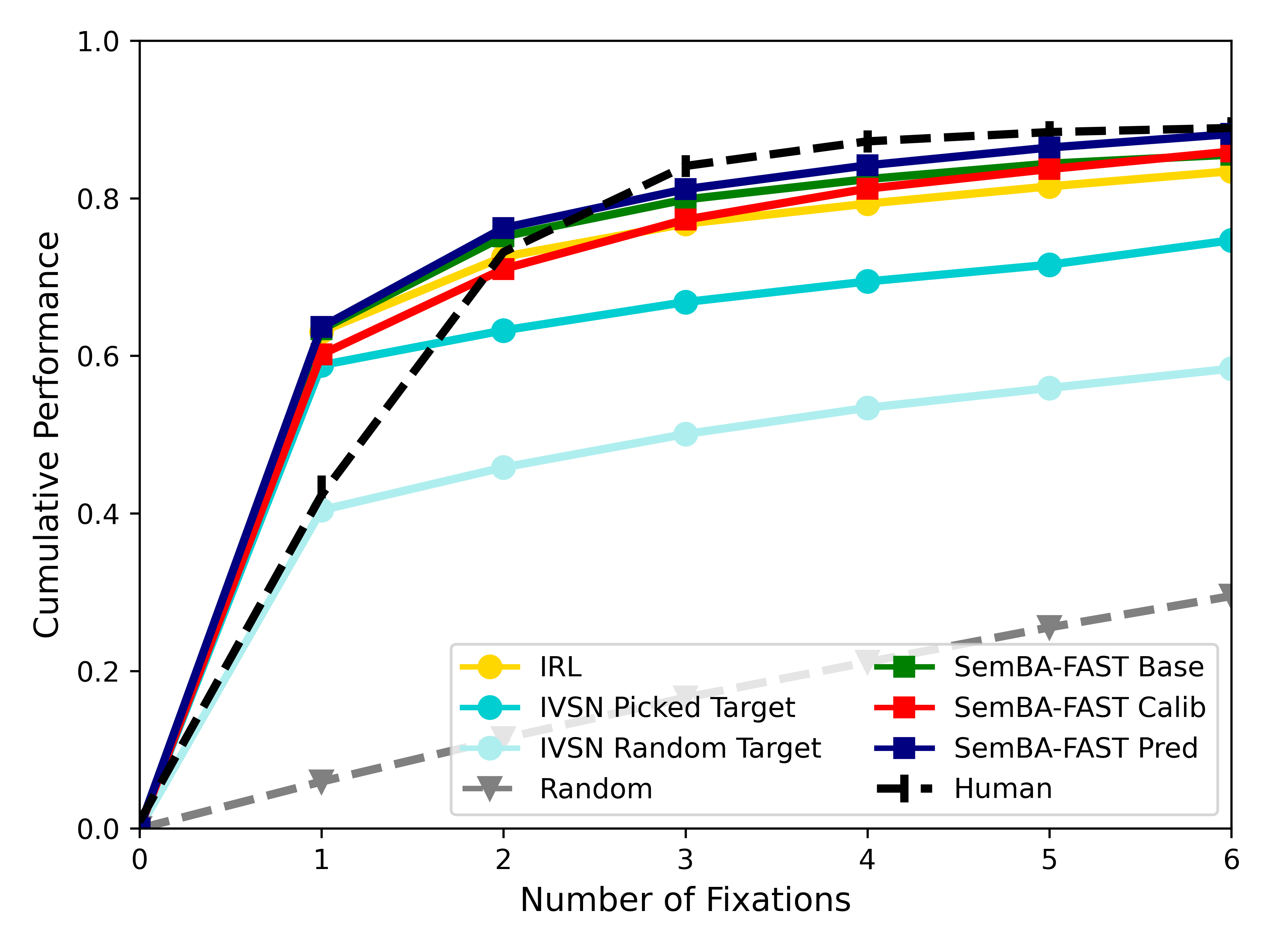}
\caption{Comparison of the cumulative performances of IRL \cite{irl}, IVSN \cite{ivsn}, and \textit{SemBA}-FAST \cite{ours} approaches, together with the average performances of humans (with standard error of the mean) and random selection.}
\label{fig:models}
\end{figure}

To properly and accurately assess the performance of \textit{SemBA}-FAST, we compare baseline and state-of-the-art models, using the previously listed metrics. First, we determine \textbf{Human Consistency}, comparing the fixations generated by each subject with its peers' scanpaths. Then, as an essential baseline, we define a \textbf{Random Fixation Selection} algorithm that consists of an aleatoric selection of fixations with IOR.    

Since IVSN \cite{ivsn} processes both the scene and target in parallel, we decided to test the impact of the target images, which are fed to the model, on the output attention maps and respective scanpaths.
As our
first
IVSN-based experiment, named \textbf{IVSN Picked Target}, we picked one particular target image, with a white background, for each of the 18 classes from COCO-Search18 \cite{cocosearch18}. 
Then, as a second
experiment, referred to as \textbf{IVSN Random Target}, we fed the model with the cropped and resized bounding-box content, extracted from a randomly selected COCO 2017 \cite{coco} image (excluding the selected scene), containing an instance of the target class.

Regarding models that depend on both visual stimuli and human fixation data, we select IRL \cite{irl}, Chen et al. \cite{vqa}, FFMs \cite{ffm}, Gazeformer \cite{gazeformer}, HAT \cite{transformers}, and CLIPgaze \cite{clipgaze}, trained on COCO-Search18 \cite{cocosearch18}. To validate our model evaluation process, we successfully replicated IRL's performance, achieving scanpath metric results (presented in Tab. \ref{tab:main}) similar to those reported in \cite{gazeformer,transformers,clipgaze}. For the remaining models \cite{ffm,vqa,transformers,gazeformer,clipgaze}, we report the scanpath metric results according to the information available on \cite{clipgaze}. Regarding the next fixation metrics, we consider and compare the performances reported by \cite{transformers} for the models \cite{irl,ffm,vqa}. 

We report the results for the assessment of \textit{SemBA}-FAST and the other mentioned scanpath prediction models in Tab. \ref{tab:main}, applying the evaluation metrics enumerated in Section \ref{sec:metrics}. We split the content presented in Tab. \ref{tab:main} into four groups, which are (from top to bottom) human and random baselines, pure stimulus-based models \cite{ivsn}, human data-dependent models \cite{irl,ffm,vqa,transformers,gazeformer,clipgaze}, and \textit{SemBA}-FAST \cite{ours} approaches. We highlight the best result achieved for each evaluation metric.

\subsection{Discussion}


Although \textit{SemBA}-FAST does not outperform models trained with human fixations \cite{ffm,vqa,transformers,gazeformer,clipgaze,irl} on all metrics, it competes favorably and even outperforms some of them on certain relevant metrics. As demonstrated by the SS, FED, SemSS and SemFED results, \textit{SemBA}-FAST's variants outperform FFMs \cite{ffm} in fixation sequence similarity while beating IRL \cite{irl} when it comes to FED and SemFED. This results highlight the potential of semantic-guided bio-inspired frameworks for human attention modelling.


Within \textit{SemBA}-FAST's variants, the attained results show that \textit{SemBA}-FAST \textit{Pred} yields the best performance. Nevertheless, \textit{SemBA}-FAST \textit{Base} does not fall that much short behind \textit{SemBA}-FAST \textit{Pred}, achieving very similar results with much less added computational costs. Startlingly, \textit{SemBA}-FAST \textit{Calib} generates the worst results among its siblings. These results suggest that, by embracing the uncertainty that naturally derives from peripheral distortion and is reflected on the detector's output semantic scores, \textit{SemBA}-FAST can more accurately predict human attention and gaze patterns.

In Fig. \ref{fig:models} we compare the cumulative performances of SemBA-FAST's approaches with IRL \cite{irl} and IVSN \cite{ivsn}, as well as the baselines. The attained results reinforce the point that \textit{SemBA}-FAST, as a pure semantic-based methodology, closely mimics human performance in TP visual search tasks, while achieving the highest performance among competitors.

Overall, the different \textit{SemBA}-FAST variants not only outperform IVSN (a top-down zero-shot method) but also demonstrate competitive performance against models trained with human scanpath data. Notably, \textit{SemBA}-FAST's approaches achieve cNSS values comparable to most of the state-of-the-art models present in Tab. \ref{tab:main}, notably surpassing IRL \cite{irl} and IVSN \cite{ivsn}. Interestingly, only FFMs \cite{ffm} and HAT \cite{transformers}, two models trained with human scanpaths, and our own \textit{SemBA}-FAST were able to achieve positive cIG values. It is important to note that we considered the human average fixation density \cite{transformers} as the baseline distribution. Surprisingly, as a model agnostic to human fixation data, \textit{SemBA}-FAST not only outperformed IRL \cite{irl} in all next fixation metrics, but also delivered cAUC values above the $90\%$ threshold.

In terms of time costs, which range between \SI{500}{\milli\second} and \SI{800}{\milli\second} per iteration for \textit{SemBA}-FAST \textit{Base} and \textit{Pred} respectively, we show that our methodology, although slower than other models, is still suitable for real-time applications. 




\section{Conclusion}

Research on human attention and goal-directed scanpath prediction \cite{sota} has been directed toward training with actual human fixation data, exploiting recently established benchmarks and datasets \cite{cocosearch18}. In this work, we show that our semantic-foveal probabilistic framework for visual search, \textit{SemBA}-FAST \cite{ours}, can approximate human performance in terms of scanpath similarity and search accuracy without learning directly from the scanpaths of actual humans. By applying an explicit foveation system \cite{fovsys}, instead of implicit retinal mechanisms \cite{ffm,transformers,gazeformer}, our model generates fixation sequences that can be visually understood in the light of the uncertainty imposed by peripheral distortion in the field of view. Moreover, we show that the distinct variants of \textit{SemBA}-FAST can compete and even outperform state-of-the-art models \cite{ivsn,irl,ffm,vqa,transformers,gazeformer,clipgaze} in some fixation prediction metrics. 

Similarly to human beings, our framework \cite{main} exploits the temporal fusion of top-down semantic information, sequentially extracted from each point of view, instead of learning from human gaze behavior. \textit{SemBA}-FAST's performance on TP visual search settings demonstrates that it is possible to model human attention by turning back to more biologically plausible fundamentals and principles \cite{wolfe} instead of following the currently established gaze-informed learning trend. 


As directions for future work, we propose evaluating \textit{SemBA}-FAST under target-absent (TA) settings \cite{ffm}, where objects of interest are not present in the scene, making them undetectable. Our framework's modular structure opens opportunities for adapting the active perception mechanism to other tasks, such as free-viewing or VQA. Studies involving the modification of the detection block \cite{detr} or the foveal geometry configuration \cite{activevision} may also yield valuable insights into the importance and limitations of each component.


\section*{Acknowledgments}

This work was supported by \textit{Fundação para a Ciência e Tecnologia} (FCT), under the LARSyS FCT funding (DOI: 10.54499/LA/P/0083/2020, 10.54499/UIDP/50009/2020 and 10.54499/UIDB/50009/2020) and the HAVATAR project (DOI: 10.54499/PTDC/EEIROB/1155/2020). João Luzio is supported by the FCT doctoral grant [2024.00683.BD].


\end{document}